\documentclass[conference]{IEEEtran}
\usepackage{cite}
\usepackage{amsmath,amssymb,amsfonts}
\usepackage{algorithmic}
\usepackage{graphicx}
\usepackage{svg}
\usepackage{textcomp}
\usepackage{xcolor}
\usepackage{siunitx}
\usepackage{comment}
\usepackage[hidelinks]{hyperref}
\usepackage{orcidlink}
\usepackage{subcaption}
\usepackage[linesnumbered,ruled,vlined]{algorithm2e}
\UseRawInputEncoding %changeit later
\def\BibTeX{{\rm B\kern-.05em{\sc i\kern-.025em b}\kern-.08em
    T\kern-.1667em\lower.7ex\hbox{E}\kern-.125emX}}

\newcommand{\todo}[1]{\textbf{\textcolor{red}{TODO: #1}}}
\usepackage{xspace} 
\newcommand{\frcpp}{\emph{FaRe-CPP}\xspace}

\usepackage{varwidth}
\DeclareCaptionFormat{myformat}{%
  \begin{varwidth}{\linewidth}%
    \centering
    #1#2#3%
  \end{varwidth}%
}

\definecolor{lightgray}{gray}{0.6}
\definecolor{darkgray}{gray}{0.4}
    
\begin{document}

\title{Fast-Revisit Coverage Path Planning\\for Autonomous Mobile Patrol Robots\\Using Long-Range Sensor Information \\

}

\author{
\IEEEauthorblockN{Srinivas Kachavarapu}
\IEEEauthorblockA{\textit{Human-Centered Robotics Lab} \\
\textit{Ostfalia University of Applied Sciences}\\
Wolfenbuettel, Germany \\
\orcidlink{0009-0002-4495-8539} 0009-0002-4495-8539
}
~\\
\and
\IEEEauthorblockN{Tobias Doernbach}
\IEEEauthorblockA{\textit{Human-Centered Robotics Lab} \\
\textit{Ostfalia University of Applied Sciences}\\
Wolfenbuettel, Germany \\
\orcidlink{0000-0001-6488-8211} 0000-0001-6488-8211
}
~\\
\and
\IEEEauthorblockN{Reinhard Gerndt}
\IEEEauthorblockA{\textit{Human-Centered Robotics Lab} \\
\textit{Ostfalia University of Applied Sciences}\\
Wolfenbuettel, Germany \\
\orcidlink{0000-0001-6415-9481} 0000-0001-6415-9481
}
}

\maketitle

\begin{abstract}
The utilization of Unmanned Ground Vehicles (UGVs) for patrolling industrial sites has expanded significantly. These UGVs typically are equipped with perception systems, e.g., computer vision, with limited range due to sensor limitations or site topology. High-level control of the UGVs requires Coverage Path Planning (CPP) algorithms that navigate all relevant waypoints and promptly start the next cycle.

In this paper, we propose the novel \emph{Fast-Revisit Coverage Path Planning} (\frcpp) algorithm using a greedy heuristic approach to propose waypoints for maximum coverage area and a random search-based path optimization technique to obtain a path along the proposed waypoints with minimum revisit time. We evaluated the algorithm in a simulated environment using Gazebo and a camera-equipped TurtleBot3 against a number of existing algorithms. 
Compared to their average path lengths and revisit times, our \frcpp algorithm showed a reduction of at least 21\% and 33\%, respectively, in these highly relevant performance indicators.
\end{abstract}

\vspace{\baselineskip} % i added to main space b%w abstract and keywords

\begin{IEEEkeywords}
%CPP - Coverage Path Planning, GRASP - Greedy Random Adaptive Search Procedure, WPD - waypoint Dropout, WPO - waypoint Optimization. 
% only allowed from this list: https://www.ieee-ras.org/conferences-workshops/fully-sponsored/icra/keywords
Motion and Path Planning, Reactive and Sensor-Based Planning, Vision-Based Navigation, Surveillance Systems, Industrial Robots
\end{IEEEkeywords}

\section{Introduction}

Robots autonomously patrolling environments like industrial, manufacturing, construction or agricultural sites are supposed to conduct efficient surveillance by covering a maximum area in minimum time. In order to ensure this, a number of Coverage Path Planning (CPP) approaches exist that compute paths passing through all predefined or computed waypoints while covering the area of interest in the least time and with the least spatial effort. Robots rely on CPP to ensure efficient coverage, which is essential for optimal performance and safety in many applications. These algorithms can be categorized into online and offline approaches \cite{Galceran2013} where online approaches generally assume little to no prior knowledge about the environment and often use heuristics to ensure maximum coverage. Offline approaches utilize pre-recorded maps of the environment, making them well-suited for large-scale areas. By leveraging hardware capabilities, all these approaches can ensure an efficient path that maximizes area coverage. Furthermore, CPP algorithms are categorized based on various factors, including the environment type (indoor, outdoor), dimensionality (2D occupancy maps, 3D occupancy maps), and the platform (unmanned ground (UGV) or aerial (UAV) vehicles). Additionally, multi-robot CPP approaches exist \cite{Almadhoun2019} which are more efficient for large areas due to task division and faster coverage, but require complex coordination and may not be applicable in all scenarios.

\begin{figure}[tb]
  \centering
  \includegraphics[width=0.9\linewidth]{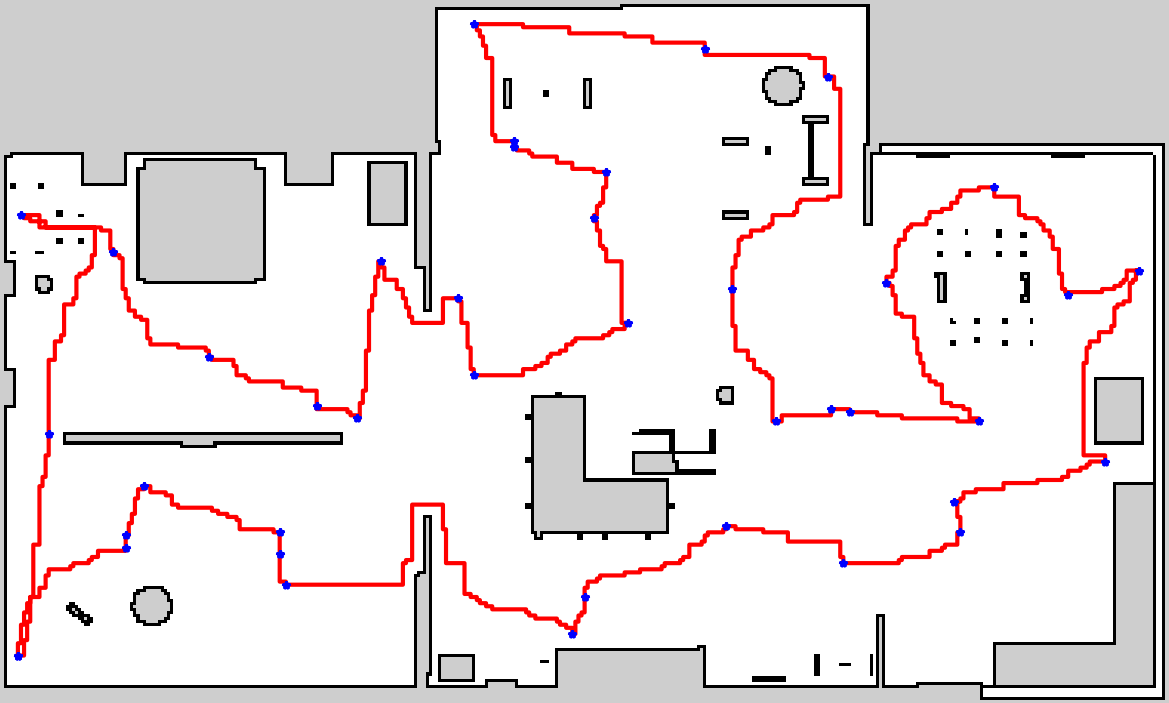}
  \caption{Proposed \frcpp algorithm example path (\textcolor{red}{red}) featuring minimum revisit times and maximum sensor coverage area}
  \label{fig:FaRe_CPP}
\end{figure}

Ensuring minimum revisit times of any given waypoint is crucial for certain application, for instance, safety patrol on hazardous substance sites like garbage dumps and recycling centers. Typically, in such sites, spontaneous fires are an imminent hazard because of inflammable substances or self-ignition because of the heat produced during composting. Therefore, fire incidents happen frequently \cite{Bahamid2020,Dhiman2021} and there is great interest by waste disposal enterprises to prevent such incidents. A safety patrol robot that is able to detect fires autonomously, and potentially even applies extinguishing agents directly, calls the fire department and keeps the fire at bay until their arrival. For such kind of patrols, minimum
revisit times with maximum coverage are essential to prevent fires from getting out of control.

In this paper, we present a novel Coverage Path Planning algorithm with focus on fast revisit times for a single mobile patrol robot. We name the algorithm \emph{Fast-Revisit Coverage Path Planning} (\frcpp).
Utilizing a pre-recorded 2D occupancy grid map, our approach updates the surveilled region within the field of view of the relevant onboard sensors at each step, subsequently creating waypoints that maximize the coverage area. Initially generated with a greedy heuristic for maximum perception potential, these waypoints are optimized using the GRASP method \cite{Feo1995GRASP} to minimize path length while preserving coverage information.
Extending existing work from the literature, our approach ensures minimal revisit times as its key contribution. This is achieved by a combination of waypoint generation and optimization which results in maximum overall area coverage with no path overlap and minimal path length. 

\textbf{A Python implementation of \frcpp} with ROS\footnote{Robot Operating System, \url{https://ros.org}} connectivity for rapid deployment in real-life scenarios \textbf{is publicly available on \url{https://github.com/hcr-lab/fare-cpp}}.

\section{Related Work}

\subsection{Basic Coverage Path Planning Methods and Applications}
Coverage Path Planning (CPP) is essential for robot applications like vacuum cleaning \cite{12333}, radiation monitoring \cite{AbdRahman2022}, bridge inspection \cite{Phung2017EnhancedDPSO}, emergency evacuation \cite{Bahamid2020}, and construction monitoring \cite{Becker2022}.

Earlier researchers implemented decomposition approaches for CPP, including exact cellular \cite{Acar2002,LI2011876}, trapezoidal \cite{Oksanen2009}, Boustrophedon \cite{Choset1998}, and Morse-based \cite{AcarChoset2002} decomposition which divides free space into navigable cells, optimizing systematic path coverage through various cell partitioning and merging techniques. Miao et al.~\cite{Miao2018} propose a scalable CPP method for cleaning robots using rectangular map decomposition, simplifying the planning process for large-scale applications. These decomposition methods focus on the robot's physical footprint rather than sensor ranges, such as camera fields of view, which are vital for scanning environments and therefore an essential part of \frcpp. Decomposition methods create zig-zag patterns and are less suited for patrolling robots, as they might disrupt ongoing activities. Gabriely et al.~\cite{Gabriely2001} discuss the Spanning Tree Covering algorithm, one of the first grid-based algorithms for CPP.

Latest methods in Coverage Path Planning move beyond merely partitioning areas and integrate both operational aspects and sensor capabilities into the area decomposition process~\cite{Swain2023,Mansouri2018,Ghaddar2020,Phung2017EnhancedDPSO,BarasDasygenis2023,TranGarrattKasmarikAnavatti2023,TranGarrattKasmarikAnavattiAbpeikar2022,KantarosThanouTzes2015,Tran2023multipleugvs}. For instance, Swain et al.~\cite{Swain2023} introduced an efficient path-planning algorithm for 2D ground area coverage using multi-UAVs, emphasizing the integration of sensor capabilities into the planning process. Mansouri et al.~\cite{Mansouri2018} developed a method for 2D visual area coverage and path planning that couples the robot's path with the camera footprints like our approach, enhancing the fidelity of CPP by considering visual sensor constraints. Similarly, Ghaddar et al.~\cite{Ghaddar2020} presented an energy-aware grid-based CPP for UAVs, incorporating area partitioning in the presence of no-fly zones, which highlights the importance of operational constraints in path planning. Phung et al.~\cite{Phung2017EnhancedDPSO} proposed an enhanced discrete particle swarm optimization algorithm for UAV vision-based surface inspection, demonstrating the benefits of considering sensor footprints for efficient inspection tasks. Kantaros et al.~\cite{KantarosThanouTzes2015} explored distributed coverage control for concave areas, considering visibility sensing constraints, further highlighting the need to integrate sensor capabilities into the CPP process. 
An et al.~\cite{An2018} introduced the Rainbow Coverage Path Planning (RCPP) method for patrolling mobile robots with circular sensing ranges, optimizing the path to minimize overlap in sensing regions and enhance patrolling efficiency.

However, none of the mentioned methods considers fast revisit times for time-critical patrol operations as is the focus of our approach.

\subsection{Established Practical Implementations of Coverage Path Planning}

Building upon the significant advancements in Coverage Path Planning techniques, certain studies have not only concentrated on theoretical models but have also ventured into practical implementations and analyses through software tools. The early work of Bormann et al.~\cite{FraunhoferIPA2016} provides an exploration of segmented rooms with a set of suitable map division algorithms. Following this foundational work, they presented a comprehensive survey alongside the implementation and analysis of various indoor CPP methods \cite{FraunhoferIPA2018}. We are using several of the analyzed methods as baseline for our approach, as described in detail in Section~\ref{results_comparison}.

%The exploration of unmanned ground vehicle (UGV) full coverage planning by Lee et al.~\cite{Lee2023} delves into the intricacies of planning in environments with negative obstacles, a crucial consideration for achieving comprehensive coverage in challenging environments. \todo{@Srini: Is Lee2023 available as a tool? Otherwise this doesn't really belong here. The last paragraph seems to be a little confusing/disconnected... They actually used these algorithms \cite{FraunhoferIPA2018}, slightly modifying them for their use case, and the tool is publically available on Git Hub. It's relevant, but it should be moved to section B, maybe (I can do that).  Yes please :-) }

\subsection{Coverage Path Planning for Environmental Operations}

Tran et al.~\cite{TranGarrattKasmarikAnavattiAbpeikar2022,TranGarrattKasmarikAnavatti2023,Tran2023multipleugvs} contributed to the dynamic and robust multi-robot coverage of both known and unknown environments through frontier-led swarming techniques. These methods underscore the importance of dynamic adaptation and sensor integration in CPP like our camera-based approach. Talavera et al.~\cite{FernandezTalavera2023} developed an autonomous ground robot to support firefighters in indoor emergencies. This robot utilizes advanced CPP algorithms to navigate complex indoor environments, detect fires, and provide real-time situational awareness to firefighting teams, demonstrating the practical utility of CPP in critical firefighting applications like we target with our work.
Nasirian et al.~\cite{Nasirian2021} addressed the challenge of effectively covering and disinfecting an environment with mobile robots using a graph-based representation like we use in our approach.
Similarly, Pierson et al.~\cite{Pierson2021} focused on deploying a mobile UVC disinfection robot with an optimized CPP algorithm. 
Perminov et al.~\cite{perminov2023} presented a genetic-based human-aware CPP algorithm for autonomous disinfection robots, achieving significant coverage efficiency.
These hygiene-optimizing robot applications are similar to our firefighting approach in the sense that they profit from fast revisit times and are deployed on robots that are to treat the environment.
%Chen et al.~\cite{Chen2024} utilize building information modeling (BIM) to enhance the efficiency of CPP for indoor robots, considering various robotic configurations for adaptable planning. This work builds on previous research~\cite{Chen2022} which explored global path planning based on BIM and physics engines for UGVs in indoor environments. Full coverage planning by Lee et al.~\cite{Lee2023} researches intricacies of path planning in environments with negative obstacles.

\section{Proposed Approach: Fast-Revisit Coverage Path Planning (\frcpp)}

\begin{figure}[t]
  \centering
  \includegraphics[width=0.85\linewidth]{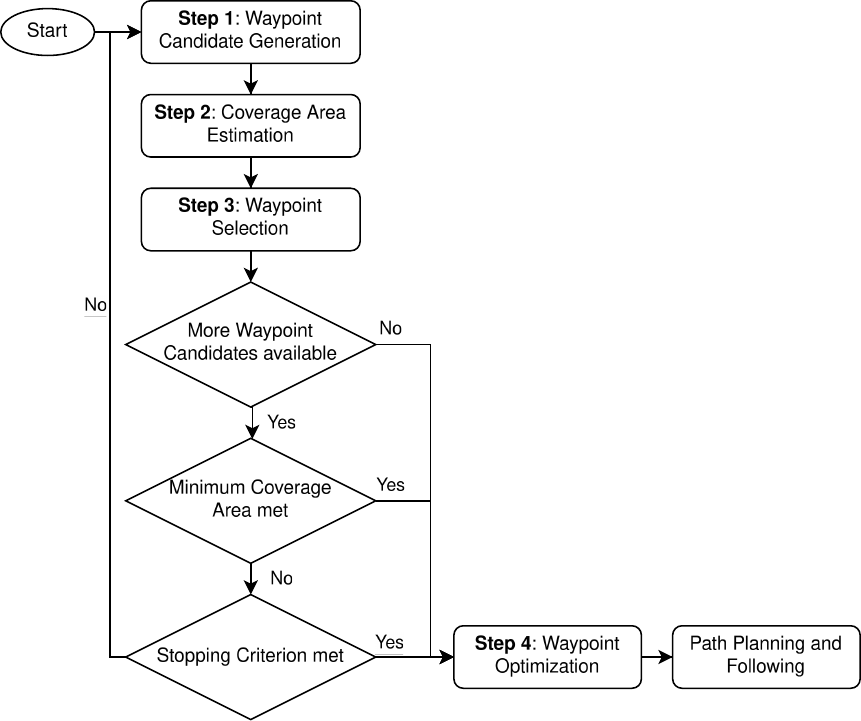}
  %Bilder/flow_chart_cpp.png
  \caption{Fast-Revisit Coverage Path Planning (\frcpp)}
  \label{fig:flow_chart}
\end{figure}

\begin{comment}
\begin{algorithm}
\DontPrintSemicolon
\KwData{grid map, initial robot pose}
\While{waypoint candidates are available \textbf{or} overall coverage area not achieved \textbf{or} stopping criterion not met}{
    generate waypoint candidates $W_c$\;
    \ForAll{$w_c \in W_c$}{
       estimate coverage area $A(w_c)$\;
    }
    select next waypoint $w = \max_{w_c \in W} A(w_c)\rightarrow w_c$
}
\Return{path}\;
\caption{\frcpp \todo{just playing around here, sorry for the spamming}}
\label{alg:frcpp}
\end{algorithm}
\end{comment}

In this section, we discuss in detail the methodology of our novel Fast-Revisit Coverage Path Planning (\frcpp) algorithm. 
Our main assumption is that the environment is represented as an occupancy map recorded using Simultaneous Localization and Mapping (SLAM) (see Section~\ref{experimental_setup}). The grid cells in the occupancy map fall into one of three categories: occupied, unoccupied, and unknown.

In \frcpp, instead of considering the robot's physical dimensions as the footprint, we utilize the field of view (FoV) of the range sensor, e.g. 2D camera, mounted on the robot as the footprint. The range within which the sensor can perceive range information is referred to as the sensor range. The grid cells within the FoV are then referred to as explored cells, and the center cell is considered the robot's position (x, y) while the orientation ($\theta$) is obtained based on the direction of the robot relative to the coordinate system of the grid map. As visually summarized in Fig.~\ref{fig:flow_chart}, there are four main steps in \frcpp which we describe in the next subsections.

\subsection{Step 1: Waypoint Candidate Generation}

The algorithm begins with the waypoint generation, where the robot's initial position is used as the first waypoint. The robot's initial FoV is then calculated based on its orientation and sensor range. For each grid cell $(x_c, y_c)$ within the sensor range $r$, the angle $\alpha$ is determined using $\alpha = \text{atan2}(y_c - y_r, x_c - x_r)$. If $\alpha$ lies within the FoV range and is not blocked by an obstacle (checked using the Bresenham line algorithm \cite{bresenham1965}), the cell $(x_c, y_c)$ is marked as explored.

\begin{figure}[t]
  \centering
  \includegraphics[width=0.8\linewidth]{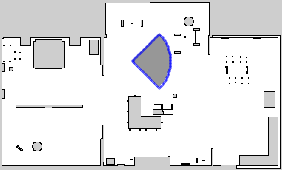}
  \caption{Explored region updated within the field of view (\textcolor{darkgray}{dark gray area}) and boundary waypoint candidates $W_c$ (\textcolor{blue}{blue line}) separating the explored from the unoccupied region}
  \label{fig:wpg at first fov}
\end{figure}

After updating the grid cells within the FoV to the explored region, the boundary cells that separate the explored region from the unoccupied region as shown in Fig.~\ref{fig:wpg at first fov} serve as waypoint candidates \( W_c = \{w_{c1}, w_{c2}, \ldots, w_{cn}\} \). These waypoint candidates represent possible future positions for the robot from which the next actual waypoint is selected Step 2.

\subsection{Step 2: Coverage Area Estimation per Waypoint Candidate}

After generating all waypoint candidates using the waypoint generation step, the coverage area $A(w_c)$ is estimated for each waypoint candidate $w_c \in W_c$:

\begin{equation}
A(w_c) = \left( \sum_{i,j} \left[ G(w_c)[i, j] = \text{explored cell} \right] \right) \times r^2,
\label{eq:coverage_area_eq}
\end{equation}

where $i$ and $j$ represent the cell indices in the sub grid map $G(w_c)$, $r$ represents the side length of each quadratic cell so that $r^2$ indicates the physical area of each cell, and $A(w_c)$ represents the explored area of each sub-grid map with the respective waypoint candidate selected as next actual waypoint.

\begin{figure}[t]
  \centering
  \begin{subfigure}[b]{0.45\columnwidth}
    \includegraphics[width=\linewidth]{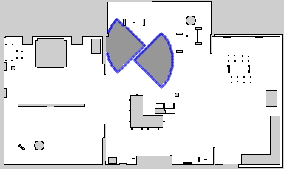}
    \caption{Sub grid map $G(w_{c1})$}
    \label{fig:wpg_first_fov_a}
  \end{subfigure}
  \hfill
  \begin{subfigure}[b]{0.45\columnwidth}
    \includegraphics[width=\linewidth]{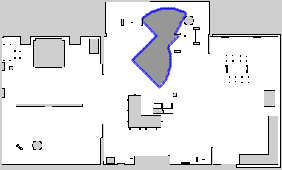}
    \caption{Sub grid map $G(w_{c2})$}
    \label{fig:wpg_first_fov_b}
  \end{subfigure}

  \vskip\baselineskip

  \begin{subfigure}[b]{0.45\columnwidth}
    \includegraphics[width=\linewidth]{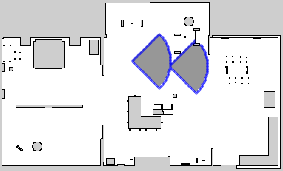}
    \caption{Sub grid map $G(w_{c3})$}
    \label{fig:wpg_first_fov_c}
  \end{subfigure}
  \hfill
  \begin{subfigure}[b]{0.45\columnwidth}
    \includegraphics[width=\linewidth]{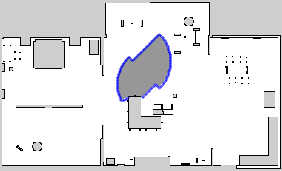}
    \caption{Sub grid map $G(w_{c4})$}
    \label{fig:wpg_first_fov_d}
  \end{subfigure}

  \caption{Exemplary sub grid maps generated for four of the boundary waypoint candidates $\{w_{c1}, w_{c2}, w_{c3}, w_{c4}\} \in W_c$ from Fig.~\ref{fig:wpg at first fov}}
  \label{fig:wpg_first_fov_all}
\end{figure}

Separate sub grid maps $G(w_c)$ per waypoint candidate are created, and the unoccupied region is updated to the explored region like in Fig. \ref{fig:wpg_first_fov_all} as discussed in Step 1. The coverage area for each grid map with the respective candidate as next waypoint is computed.

\subsection{Step 3: Waypoint Selection by Maximum Coverage Area}
After the coverage area has been computed for each waypoint candidate, the candidate with the potential of providing the maximum FoV is selected as the next waypoint \(w\) with 
\begin{equation}
w = \max_{w_c \in W} A(w_c)\rightarrow w_c.
\end{equation}
The sub-grid map of this waypoint is then used in the next iteration where new waypoint candidates are generated. 

%\todo{ please reconsider the colors of fig 5 - to me there are three greys (and in b/w printing there may be even five) if light grey would be the explored region (some not 'visible' to the robot due to walls), what is the white space?) moreover, it appears there is a red path (final path) - at right center - passing through white area and through a gap of wall thickness. Moreover, there is a waypoint at the edge of an unexplored (white) aree  - center bottom). This does not make sense - Question: do we consider the robot footprint in path finding??? Remark srini: in the above path currently no but it's possible to consider the robot footprint then the path would change a bit eg. the path won't go through tight spaces.  }

The \frcpp algorithm repeats Steps 1--3 until the stopping criterion is met. This is the case when no more waypoint candidates are available, a predefined minimum coverage area $A_{\text{min}}$ has been exceeded, or the rate of increase in the coverage area across iterations $\frac{\Delta A_i}{A_i}$ is less than a threshold $\epsilon$:

%\todo{@Srini: What is the difference between the two latter metrics? They sound similar to me -- it would be good if we could reduce this to one for readability. Also, the text says "second OR third" metric, so I changed the "equation" to "or" as well, is this correct?}

\[
\text{Stopping Criterion} = 
\begin{cases} 
\text{Explored Area } (A_{\text{total}}) > A_{\text{min}}, \\
\text{or} \\
%\frac{\Delta A_i}{A_i} < \epsilon \text{ and } \text{Var}\left(\{\Delta A_{i-n+1}, \ldots, \Delta A_i\}\right) < \delta 
\frac{\Delta A_i}{A_i} < \epsilon \\ 
\end{cases}
\]

Our waypoint selection strategy ensures that at every iteration, we select a waypoint globally from all possible candidates. By evaluating coverage potential over the entire set of waypoint candidates, we prevent the risk of the algorithm converging to a suboptimal local minimum.

\begin{figure}[t]
  \centering
  \includegraphics[width=0.85\linewidth]{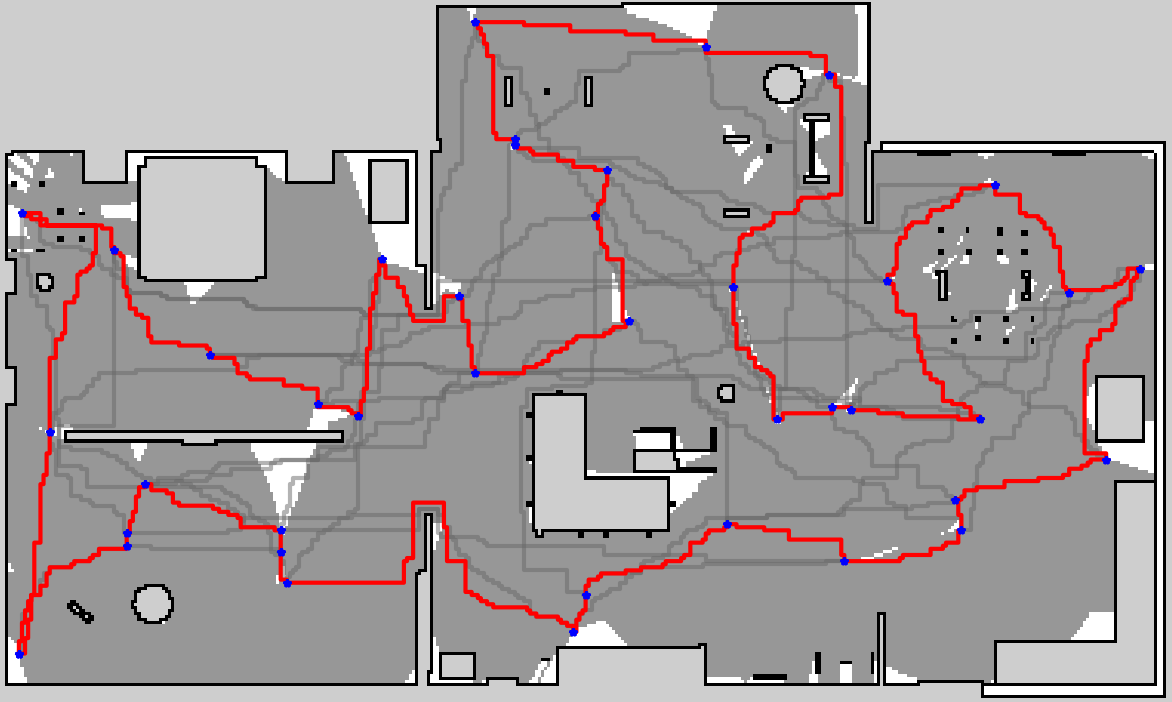}
  %{Bilder/path_before_after_wpo.png}
  %\includegraphics[width=\linewidth]{Bilder/path_before_after_wpo_2.png}
  \caption{\frcpp \textcolor{blue}{waypoints (blue)}, \textcolor{lightgray}{explored region (light gray)}, \textcolor{darkgray}{path before Waypoint Optimization (dark gray)} and \textcolor{red}{final resulting path after Waypoint Optimization (red)}}
  \label{fig:WPO}
\end{figure}

\subsection{Step 4: Waypoint Optimization}

So far, the \frcpp algorithm focuses on incrementally expanding the coverage area by selecting waypoints that maximize the coverage area at each iteration, rather than minimizing the overall path length. As a result, the waypoints are not arranged in an optimal sequence, leading to inefficient, overlapping, or zigzagging paths. This disorganization is since the waypoints are obtained in The waypoint generation step  purely based on coverage needs, not on path efficiency. To balance extensive coverage with path efficiency, we employ the Greedy Randomized Adaptive Search Procedure (GRASP) \cite{Feo1995GRASP}. GRASP operates in two phases: 1.) initial greedy randomized construction, which builds a path by selecting from top candidates, introducing randomness to avoid local optima, followed by 2.) local search, which refines the path using the 2-opt swap strategy to minimize cumulative distance.

\begin{comment}
    This can lead to overlapping paths or multiple revisits of the same areas, resulting in an inefficient path with non-optimal revisit times. To balance extensive coverage with path efficiency, we employ the Greedy Randomized Adaptive Search Procedure (GRASP) \cite{Feo1995GRASP}. GRASP operates in two phases: an initial greedy randomized construction followed by iterative local search refinements. 
\end{comment}

To avoid computational complexity, the environment map is also not considered during waypoint optimization. The optimization focuses only on ordering the waypoints, while path planning that takes the environment into account is done in the next step. GRASP optimizes the waypoints such that the global path that passes through the waypoints bears the least cumulative distance, as shown in Fig.~\ref{fig:WPO}. However, due to the introduction of randomness in the greedy construction phase and the iterative nature of the 2-opt swap strategy, there is some variability in the resulting global path. While the global path is optimized, the exact distance can vary slightly across different runs due to the stochastic elements of the algorithm. Despite this, GRASP consistently arranges waypoints with reduced cumulative distance and improved efficiency.

\subsection{Path Planning and Following}
After generating and optimizing waypoints, they are used to plan a global path. In this case, the A* algorithm \cite{hart1968astar} is employed to find a path that passes through these waypoints. A* is a graph-based search algorithm that finds the shortest path from a start point to a goal, considering the cost of movement between points. The result is a global path that connects the waypoints while maintaining alignment with the range sensor’s field of view. This path, which forms a closed loop through all waypoints, can be utilized by and path following approach to locomote the robot accordingly. In the implementation of our example scenario as used in the following experimental evaluation, we use the Dynamic Window Approach (DWA) from the ROS navigation stack for local path planning.

%\subsection{Online Path Following}
%\todo{Does our algorithm include a path follower? I don't think so?! So which "local online path planner" is used? No, we do not include any specific path follower. Instead, we use the default local path planner called the Dynamic Window Approach (DWA), which is part of the ROS navigation stack package. During the online navigation phase, we iteratively provide all our optimized waypoints, and the ROS navigation stack handles both global path planning and local path planning. I will describe this more clearly in this section.}

%The path generated by \frcpp is executed by the robot's path-following controllers, guiding it through optimized waypoints while maintaining alignment with the camera’s field of view. This path, which forms a closed loop through all waypoints, is used as a reference global path. During the online navigation phase, rather than employing a specific path follower, we utilize the Dynamic Window Approach (DWA) from the ROS navigation stack for local path planning. The robot iteratively follows the optimized waypoints, with the ROS stack managing both global and local adjustments. If unexpected obstacles or environmental changes occur, the DWA dynamically adjusts the path, ensuring continuous coverage efficiency. This process continues until the robot has visited all waypoints, maintaining the orientation of the field of view and completing the closed loop for patrolling.

\section{Evaluation}

\subsection{Experimental Setup}\label{experimental_setup}

%We conducted all the experiments in a simulation environment using the ROS Noetic framework and Gazebo. First, we created an occupancy grid map of the environment using the SLAM technique, specifically utilizing the GMapping package~\cite{gmapping2007}. We consistently use the same map which is the house world from AWS RoboMaker\footnote{\url{https://github.com/aws-robotics/aws-robomaker-small-house-world}}. This occupancy grid map has dimensions of 500x500 grid cells with a resolution of \SI{0.05}{m/cell}, resulting in an unoccupied coverage area of \SI{157}{m^2}. We chose wide-angle cameras with two different horizontal fields of view (FoV) of \ang{90} and \ang{120} and a visual radius of \SI{3.5}{m} for the following experiments.

We conducted all experiments in a simulation environment using ROS Noetic, Gazebo, and a camera-equipped TurtleBot3. We first created an occupancy grid map using the SLAM technique with the GMapping package~\cite{gmapping2007}. We selected two environments from AWS RoboMaker: The House world\footnote{\url{https://github.com/aws-robotics/aws-robomaker-small-house-world}} for comparison with algorithms designed for smaller environments and the Warehouse world\footnote{\url{https://github.com/belal-ibrahim/dynamic_logistics_warehouse}} for larger environments like warehouses or recycling centers.

The occupancy grid map of the House environment has dimensions of 500x500 grid cells with a resolution of \SI{0.05}{m/cell}, resulting in an unoccupied coverage area of \SI{157}{m^2}, whereas the Warehouse environment has 380x640 grid cells with a resolution of \SI{0.05}{m/cell} and an unoccupied coverage area of \SI{232}{m^2}.

\subsection{Evaluation Metrics}
We evaluated the performance of \frcpp on the generated grid map using several key metrics, chosen to comprehensively assess the algorithm's efficiency and effectiveness in selected environments:

\begin{itemize}
\item \textbf{Path Length} $l_{p}$: This metric measures the total distance traveled by the robot in a closed loop, starting and ending at the first waypoint. It provides insight into the efficiency of the generated path, with shorter paths being more optimal with respect to minimum revisit times.

\item \textbf{Total Cumulative Rotation} $\theta_{\mathrm{total}}$: This is calculated based on the robot's orientation changes at each waypoint. While maintaining maximum coverage, this should be minimized in order to achieve fast revisit times.

\item \textbf{Revisit Time} $t_{\mathrm{revisit}}$: Revisit time measures the duration required for the robot to complete its patrol loop and return to the starting point for subsequent patrols. It is determined as $t_{\mathrm{revisit}} = \frac{l_{p}}{v_{\mathrm{linear}}} + \frac{\theta_{\mathrm{total}}}{v_{\mathrm{angular}}}$ with $v_{\mathrm{linear}}$ as the actual linear and $v_{\mathrm{angular}}$ as the actual angular robot velocity.

\item \textbf{Coverage Area}: Percentage of the environment's floor plan covered by the sensor's field of view.

\item \textbf{Computation Time}: Total time to compute the final resulting path.

\end{itemize}

\subsection{Results}\label{results}

\begin{figure}[t]
  \centering
  \includegraphics[width=0.85\linewidth]{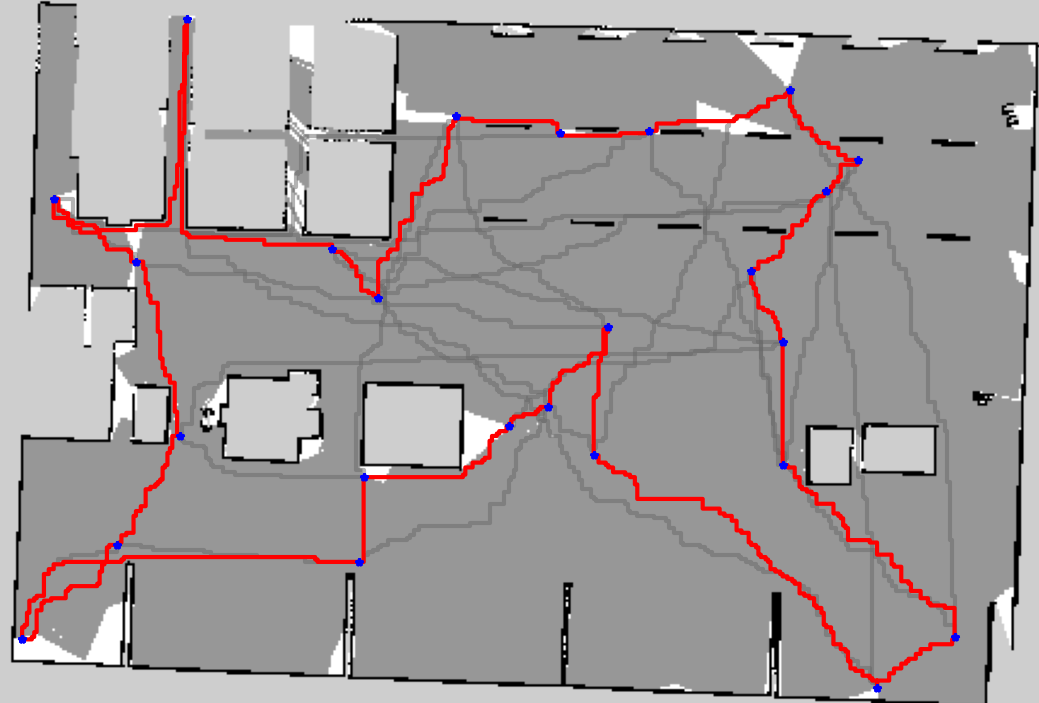}
  \caption{\frcpp on Warehouse environment -- \textcolor{blue}{waypoints (blue)}, \textcolor{lightgray}{explored region (light gray)}, \textcolor{darkgray}{path before Waypoint Optimization (dark gray)} and \textcolor{red}{final resulting path after Waypoint Optimization (red)}}
  \label{fig:FaRe_CPP_ware_house}
\end{figure}

As range sensors, we employ wide-angle cameras with FoVs of \ang{90} and \ang{120} on the Warehouse map, each with a 5-meter sensor radius. We use this radius as a conservative estimate to prove our algorithm's utility because typical low-cost sensors as used on patrol robots have a \SIrange{5}{10}{m} range.
Table~\ref{table: Comparision over cameras with FOV} shows that the \ang{120} FoV outperforms the \ang{90} FoV, with shorter computation time (\SI{149}{s} vs.\ \SI{178}{s}), better coverage (\SI{96}{\%} vs.\ \SI{94}{\%}), and reduced cumulative rotation. These results demonstrate the effectiveness of \frcpp in large environments like the Warehouse, see Fig.~\ref{fig:FaRe_CPP_ware_house}. The algorithm achieves high coverage and minimizes cumulative rotation, contributing to efficient sensor usage. Importantly, the similar revisit times (\SI{421}{s} for \ang{90} FoV and \SI{419}{s} for \ang{120} FoV) indicate that \frcpp maintains consistent performance without unnecessary re-exploration, making it well-suited for large-scale applications where extensive coverage and minimized redundancy are critical. 
%\todo{120° outperforming 90° is quite trivial unless you consider pixel size -typically a long focal length comes with a smaller angel...}

\begin{table}[tb]
\centering
\caption{\frcpp quantitative results with FoV (\ang{90},\ang{120})}
\label{table: Comparision over cameras with FOV}
\begin{tabular}{|l|c|c|}
\hline
\textbf{Metric}  & \textbf{FoV(\ang{90})} & \textbf{FoV(\ang{120})} \\ \hline
Path Length (m)              &        100.95                 &  102.15                     \\ \hline
Total Cumulative Rotation (rad) &     43.98            &  40.84                      \\ \hline
Revisit Time (s)             &            421             &    419                     \\ \hline
Coverage Area (\%)           &          94          &      96                \\ \hline
Computation Time (s)         &         178                &  149                       \\ \hline
\end{tabular}
\end{table}

\subsection{Comparison with Existing Algorithms}\label{results_comparison}

% graphs after applying WPO on all algorithms
%grid based 
\begin{figure*}[tb]
    \centering

    \begin{subfigure}{0.3\textwidth}
      \centering
      \includegraphics[width=\linewidth]{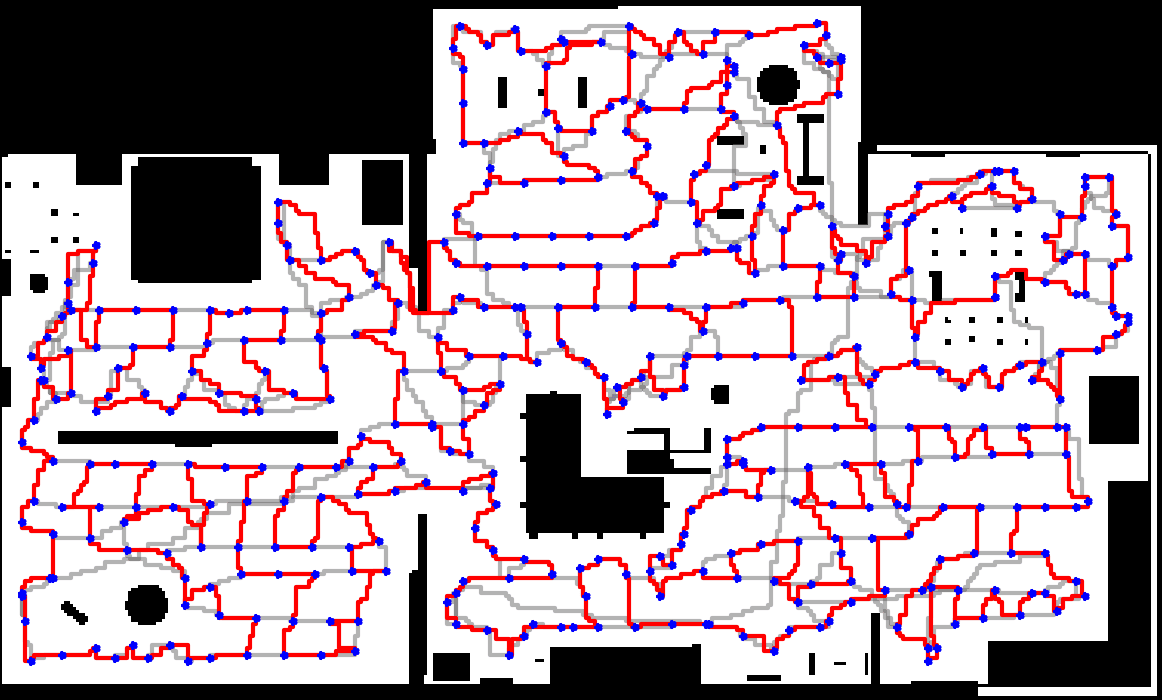}
      \caption{Grid-based TSP \cite{FraunhoferIPA2018}}
      \label{fig: Grid-based TSP}
    \end{subfigure}
    \hfill
    %Boustrophedon
    \begin{subfigure}{0.3\textwidth}
      \centering
      \includegraphics[width=\linewidth]{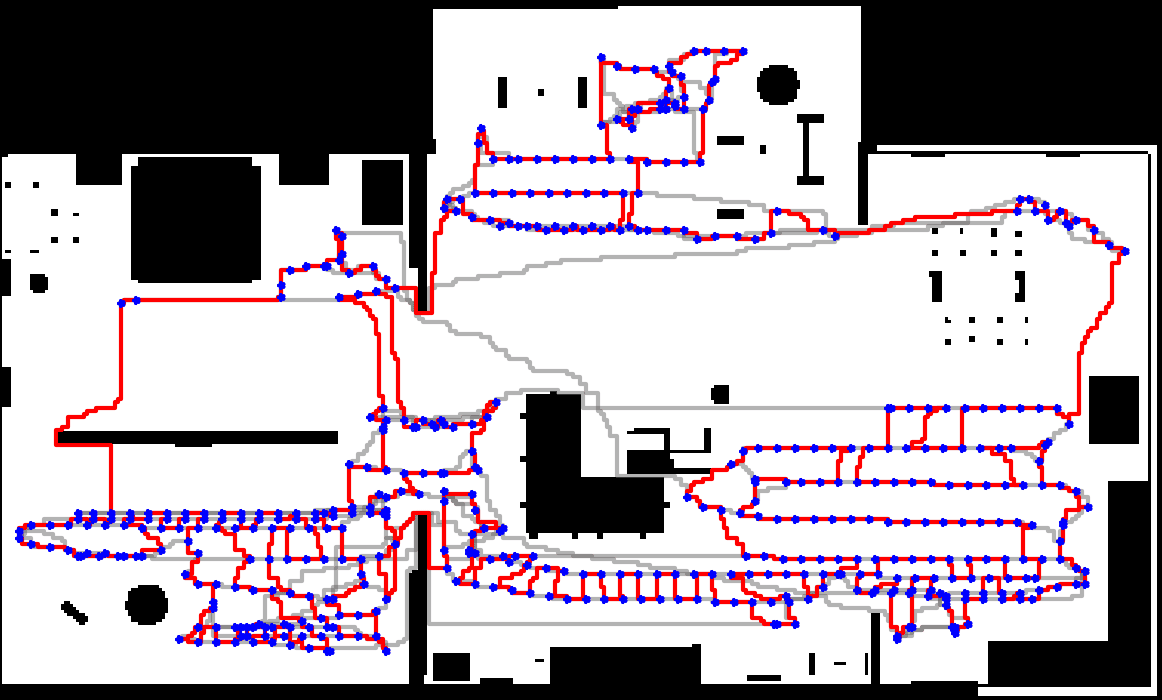}
      \caption{Boustrophedon \cite{Choset1998}}
      \label{fig: Boustrophedon}
    \end{subfigure}
    \hfill
    %Neural Network based
    \begin{subfigure}{0.3\textwidth}
      \centering
      \includegraphics[width=\linewidth]{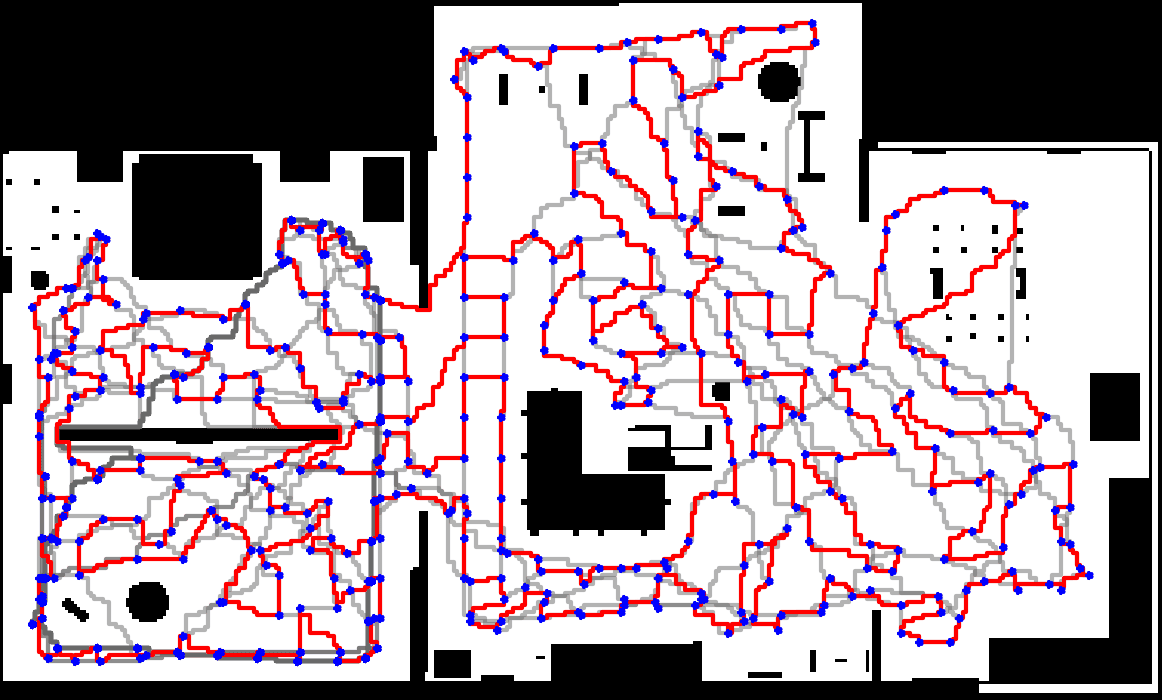}
      \caption{Neural Network-based \cite{Yang2004}}
      \label{fig:Neural Network based}
    \end{subfigure}

   \vskip\baselineskip

    %convex spp based
    \begin{subfigure}{0.3\textwidth}
      \centering
      \includegraphics[width=\linewidth]{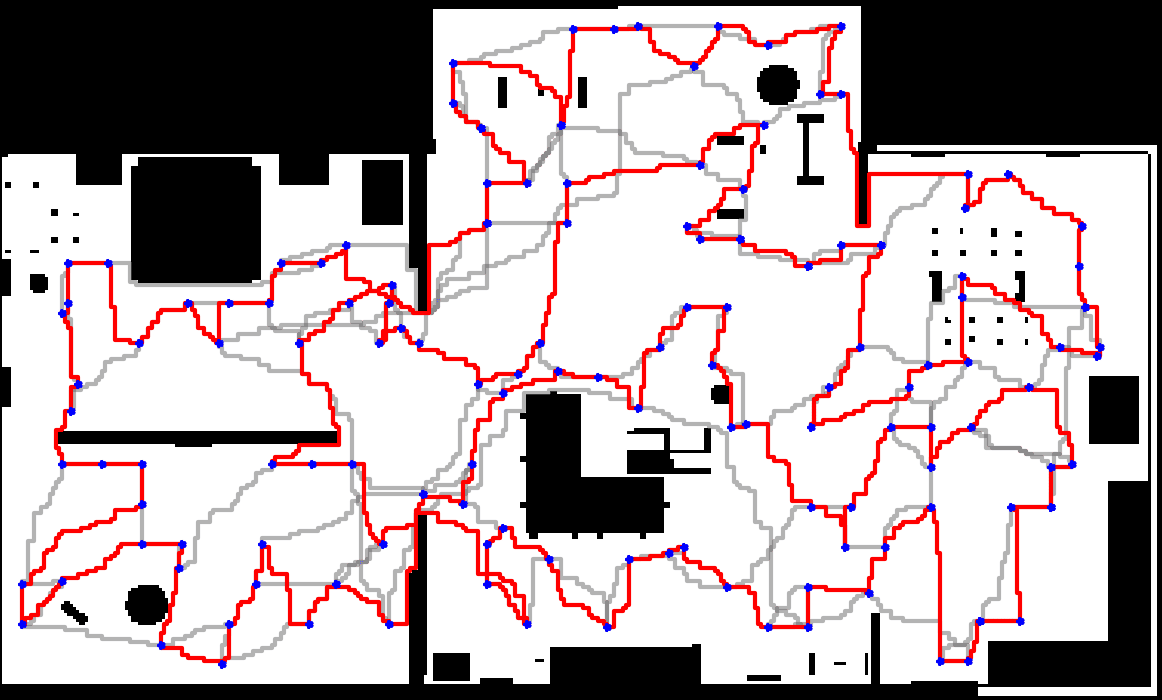}
      \caption{Convex SPP \cite{Arain2015}}
      \label{fig: Convex SPP}
    \end{subfigure}
    \hfill
    %Contour Line
    \begin{subfigure}{0.3\textwidth}
      \centering
      \includegraphics[width=\linewidth]{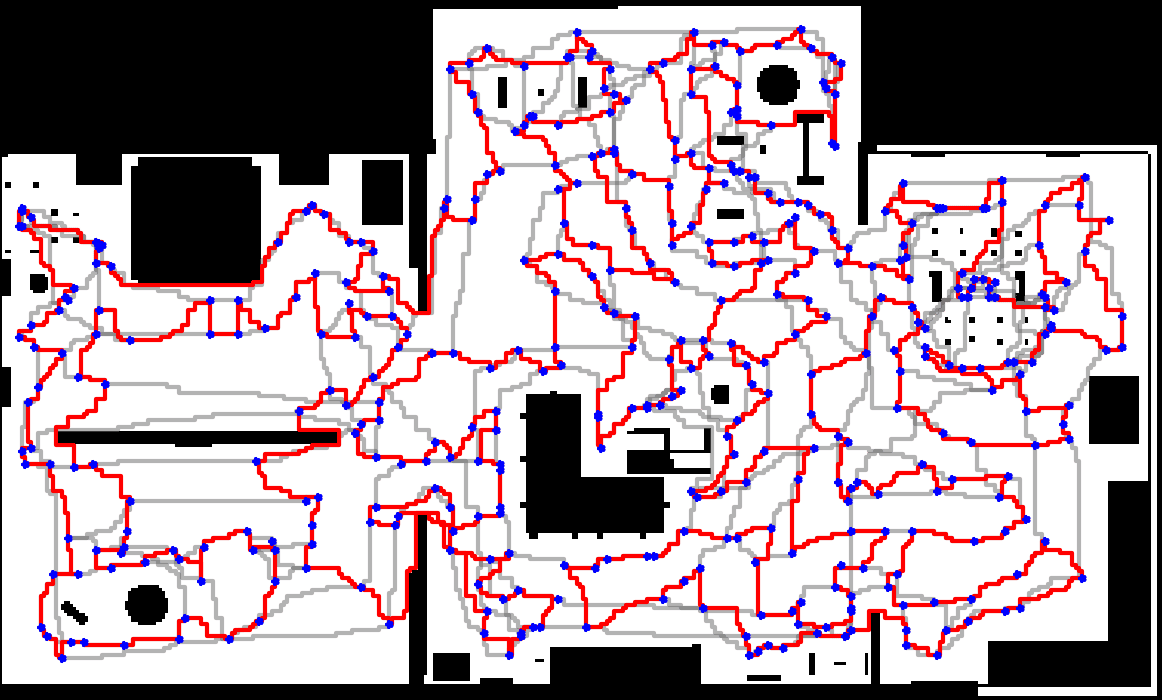}
      \caption{Contour Line \cite{FraunhoferIPA2018}}
      \label{fig: Contour Line}
    \end{subfigure}
    \hfill
    %Grid Local Energy
    \begin{subfigure}{0.3\textwidth}
      \centering
      \includegraphics[width=\linewidth]{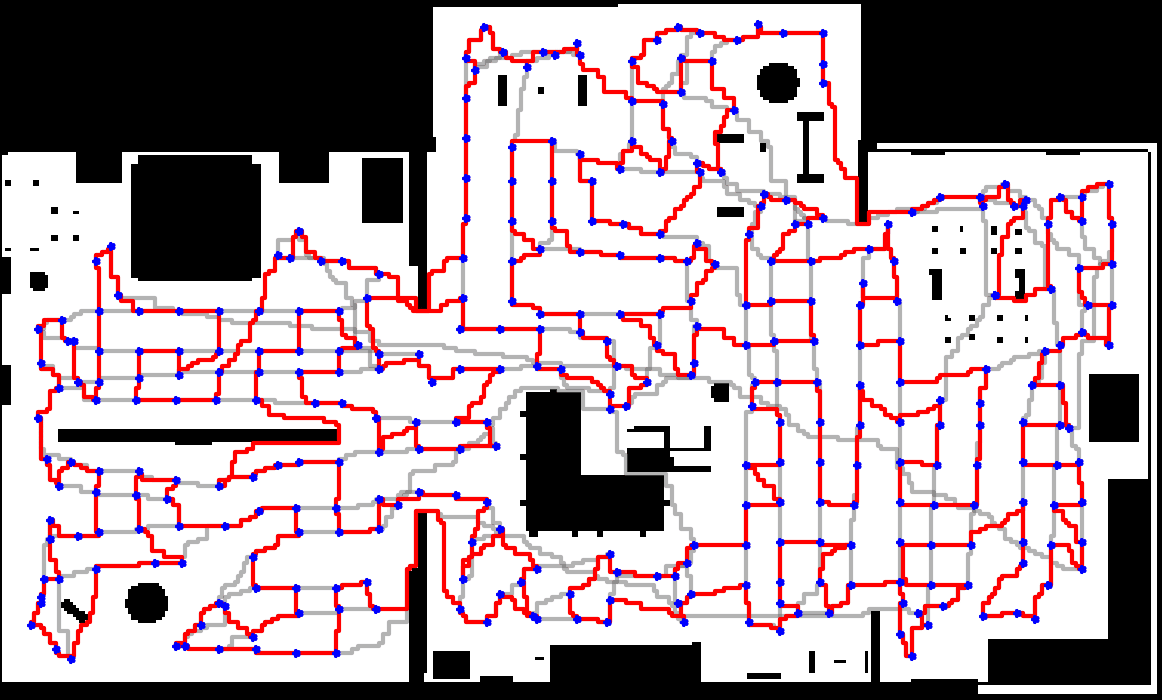}
      \caption{Grid local Energy \cite{Bormann2015}}
      \label{fig: Grid local energy}
    \end{subfigure}

   \vskip\baselineskip

   \begin{subfigure}{0.3\textwidth}
    \captionsetup{format=myformat}
      \centering
      \includegraphics[width=\linewidth]{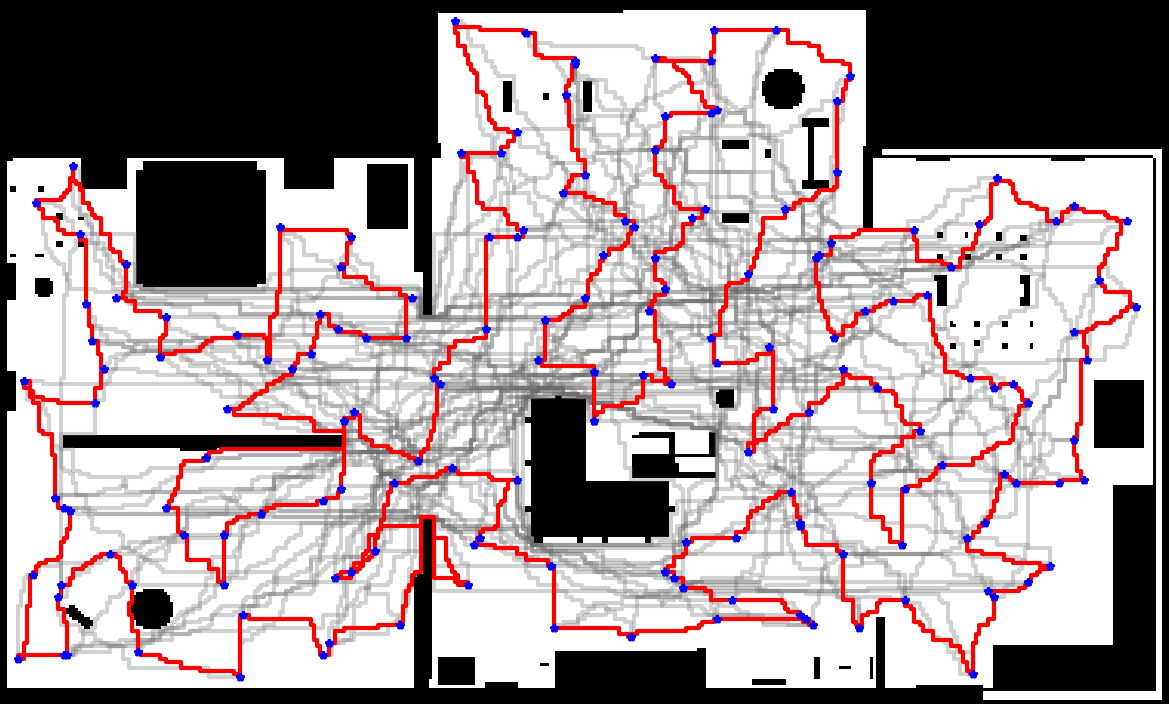}
      \caption{\textbf{Proposed \frcpp}\\(Bormann et al.~\cite{FraunhoferIPA2018} parameters)}
      \label{fig:results_our_fare_cpp_ipa_fov}
    \end{subfigure}
    %\hfill
    \hspace{1cm}    
   \begin{subfigure}{0.3\textwidth}
    \captionsetup{format=myformat}
      \centering
      \includegraphics[width=\linewidth]{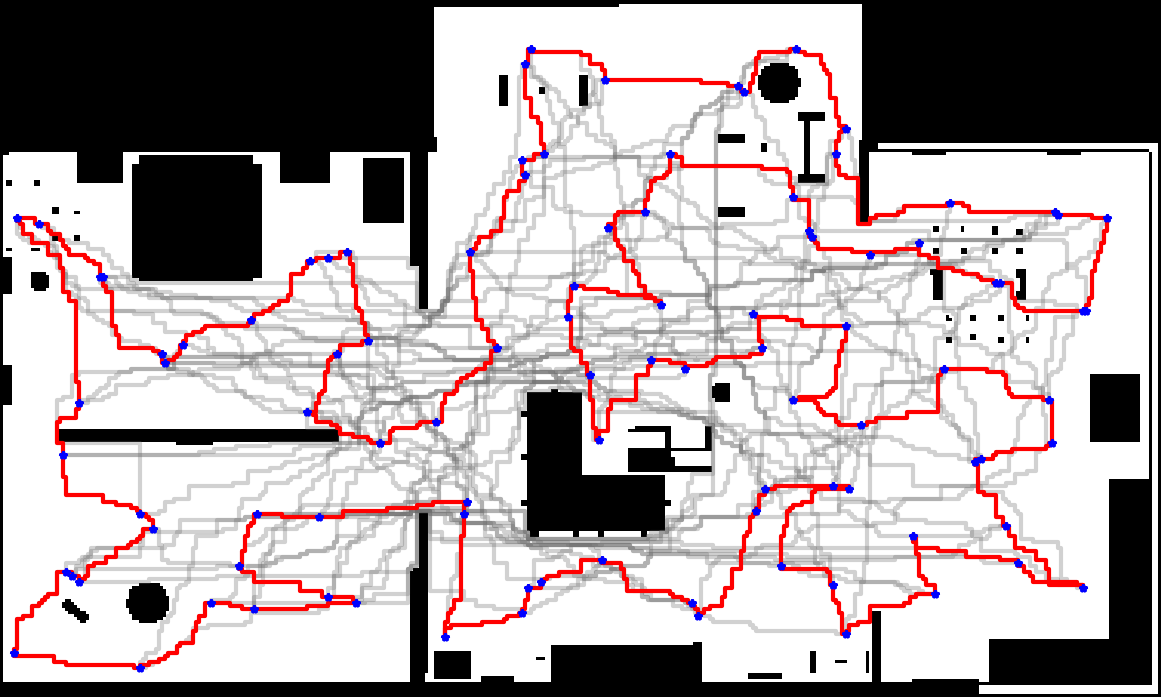}
      \caption{\textbf{Proposed \frcpp}\\(optimized for long-range efficiency)}
      \label{fig:results_ours_after_wpo}
    \end{subfigure}

\caption{Qualitative results for paths generated by existing approaches and \frcpp\ -- \textcolor{blue}{waypoints (blue)}, \textcolor{darkgray}{path before Waypoint Optimization (gray)} and \textcolor{red}{final resulting path after Waypoint Optimization (red)}}
  \label{fig:results_qual}
\end{figure*}
\begin{comment}
\begin{table*}[bt]
\centering
\caption{Quantitative average results for 10 comparison runs between existing approaches and \frcpp}

\label{table:exploration_comparison}

\begin{tabular}{|l|c|c|c|c|c|}
\hline
\textbf{Algorithm}  & \textbf{Comp.\ Time (s)} & \textbf{Path Length (m)} & \textbf{Total Rotation (rad)} & \textbf{Coverage Area (\%)} & \textbf{Revisit Time (s)} \\ \hline
Grid TSP\cite{FraunhoferIPA2018} & 148 & 297.45 & 470& \textbf{97} & 1895\\  \hline
Boustrophedon\cite{Choset1998} & 4 & 209.79 & 234& 76 & 1149\\ \hline
Neural Network\cite{Yang2004} & 8 & 293.69 & 505& 92 & 1950\\ \hline
Convex SPP\cite{Arain2015} & 17 & 171.60& 230& 92 & 1014\\ \hline
Grid Local Energy\cite{Bormann2015} & \textbf{1.7} & 261.34 & 301& 95 & 1449\\ \hline
Contour Line\cite{FraunhoferIPA2018} & 71 & 283.74 & 467& \textbf{97} & 1843\\ \hline
\textbf{Proposed \frcpp \todo{}} & 298 {$\pm 0$}& \textbf{199.65 $\pm$ 1.66}& \textbf{272 $\pm$ 0}& 92 $\pm$ 0& \textbf{1188 $\pm$ 5.53}\\
(similar parameters to other algorithms) & & & & & \\ \hline
\textbf{Proposed \frcpp} & 109 {$\pm 0$} & \textbf{148.00 $\pm$ 1.66} & \textbf{141 $\pm$ 0}& 95 $\pm$ 0 & \textbf{764 $\pm$ 5.53}\\
(optimized for long-range efficiency) & & & & & \\ \hline

\end{tabular}
\end{table*}
\end{comment}

\begin{table*}[bt]
\centering
\caption{Quantitative average results for 10 comparison runs between existing approaches and \frcpp}

\label{table:results_quant}

\begin{tabular}{|l|c|c|c|c|c|}
\hline
\textbf{Algorithm}  & \textbf{Comp.\ Time (s)} & \textbf{Path Length (m)} & \textbf{Total Rotation (rad)} & \textbf{Coverage Area (\%)} & \textbf{Revisit Time (s)} \\ \hline
Grid TSP\cite{FraunhoferIPA2018} & 148 & 297.45 & 544 & \textbf{97} & 2037 \\  \hline
Boustrophedon\cite{Choset1998} & 4 & 209.79 & 302 & 76 & 1280 \\ \hline
Neural Network\cite{Yang2004} & 8 & 293.69 & 795 & 92 & 2507 \\ \hline
Convex SPP\cite{Arain2015} & 17 & \textbf{171.60} & 347 & 92 & 1239 \\ \hline
Grid Local Energy\cite{Bormann2015} & \textbf{1.7} & 261.34 & 322 & 95 & 1490 \\ \hline
Contour Line\cite{FraunhoferIPA2018} & 71 & 283.74 & 574 & \textbf{97} & 2049 \\ \hline
\textbf{Proposed \frcpp } & 274 $\pm$ 5.89 & 199.65$\pm$ 5.06 & \textbf{272 $\pm$ 13.63}& 92 $\pm$ 0& \textbf{1188$\pm$ 28.35}\\
(Bormann et al.~\cite{FraunhoferIPA2018} parameters) & & & & & \\ \hline
\textbf{Proposed \frcpp} & 109 $\pm$ 3.26 & \textbf{148.00 $\pm$ 3.63} & \textbf{141 $\pm$ 5.91}& 95 $\pm$ 0 & \textbf{764 $\pm$ 15.29}\\
(optimized for long-range efficiency) & & & & & \\ \hline

\end{tabular}
\end{table*}

We evaluate the effectiveness of our proposed \frcpp algorithm through qualitative comparisons (Fig.~\ref{fig:results_qual}) and quantitative metrics (Table \ref{table:results_quant}) against existing camera foot print-based methods. As baselines, we utilize the algorithms evaluated in Bormann et al.~\cite{FraunhoferIPA2018} as implemented in their publicly available framework\footnote{\url{https://github.com/ipa320/ipa_coverage_planning}}, which represent the state of the art in coverage path planning. For consistency and fairness, we adopt the sensor parameters specified in their work: an RGB-D sensor mounted at 65 cm height, facing downward, with a trapezoidal field of view (FoV) with a  70 cm (closest edge, 15 cm from the robot center) to 1.3 m (farthest edge, 1.15 m from the robot center). This setup yields a horizontal FoV of \ang{58.4}, a vertical FoV of \ang{47.4}, and a sensor range of 1.3 m. In addition, we evaluate the optimal parameter set for our approach, which assumes a sensor parallel to the floor, producing a conical FoV constrained by a horizontal field of view and sensing range of 1.3 m . To enable a direct comparison, we evaluate \frcpp under these two configurations: (1) with parameters matching Bormann et al.'s~\cite{FraunhoferIPA2018} setup, and (2) optimized for long-range efficiency, our primary objective.

As the results show in Fig.~\ref{fig:results_qual}, \frcpp achieves more efficient coverage in both configurations with fewer overlaps and a more streamlined path compared to the other methods. Table \ref{table:results_quant} quantitatively supports these findings, which shows averages for 10 comparison runs. \frcpp exhibits the shortest path length, lowest total cumulative rotation, and fastest revisit time while maintaining a high coverage area percentage.

In total, compared to the average of the baseline approaches, our algorithm achieves a reduction in path length of 21\% for similar and 42\% for long range-optimized parameter settings.
% calculation of the numbers based on: 1-(avg(2037,1280,2507,1239,1490,2049)/1188) and .../746)
This results in a reduced revisit time of 33\% for similar and 57\% for long range-optimized parameter settings.
These metrics highlight the performance of \frcpp in achieving near-optimal coverage with fast revisit times for subsequent patrolling, which is an essential requirement for the targeted use cases.
\begin{comment}
With respect to computing time, \frcpp currently falls behind the other algorithms in their respectively used implementation except for Grid TSP. The main reason is that the currently available implementation of our approach uses Python where all others are implemented in C++. While \frcpp works heuristically and often finds near-optimal solutions, its computation time can be exponential, especially as the size and complexity of the grid map increase.
\end{comment}

With respect to computing time, \frcpp currently lags behind other algorithms except for Grid TSP. To ensure consistent comparison, all methods use the same GRASP-based waypoint optimization, implemented uniformly in Python. The difference lies in waypoint generation, which is implemented in Python for \frcpp and in C++ for the baselines. However, since this stage is heuristic and has minimal effect on final path quality, the impact is limited to computation time, not algorithmic efficiency. 

\begin{comment}
Thus, while we acknowledge the language-level performance differences, they do not bias the core comparison, as computation-heavy components are treated uniformly across implementations.
\end{comment}
\frcpp performs well in coverage efficiency as well as path optimization and the main issue of the application problem at hand is closed-loop revisit time after initial deployment. Since no near-realtime reprocessing of the path is required after the robot started to patrol, any reprocessing can happen during deployment while the robot is on patrol. Initial computation, in contrast, plays a comparably minor role since the main focus of our approach is to achieve fast revisit times.

We provide the experimental results described in this section as generated with the current version of \frcpp together with the code on \url{https://github.com/hcr-lab/FaRe-CPP/tree/main/experiments}.

%\todo{@Srini: please make sure this is the latest version, upload into the Github repo "experiments" subfolder with a (very short) Markdown README.md that explains what is what and do not change after Monday -- old URL: https://drive.google.com/drive/folders/1knZSJI4-RcrsCqG9sm492zueEIg6Vh-r?usp=sharing}

\section{Conclusion}

The proposed \frcpp coverage path planning algorithm achieves near-complete coverage of the environment with short paths and low total cumulative rotation, resulting in fast revisit times. Our approach effectively first generates waypoints that yield the maximum visual coverage, and subsequently optimizes the waypoints using the Greedy Randomized Adaptive Search Procedure (GRASP).

We evaluated our approach in a simulation environment, demonstrating that our algorithm outperforms existing approaches by achieving optimal coverage areas with shorter path lengths, lower cumulative rotation, and faster revisit times.
Compared to the average path lengths and revisit times of existing approaches, \frcpp showed a reduction of at least 21\% and 33\%, respectively, in these highly relevant performance indicators.

However, so far \frcpp is limited to scenarios with static obstacles and consistent coverage parameters since the path is being generated offline. In future work, we aim to expand \frcpp to be able to deal with dynamic environments with online obstacle avoidance and varying coverage requirements. This will help to further improve the robustness and adaptability of \frcpp in real-world scenarios.

\bibliographystyle{IEEEtran}
\bibliography{references} 

\begin{thebibliography}{10}
\providecommand{\url}[1]{#1}
\csname url@samestyle\endcsname
\providecommand{\newblock}{\relax}
\providecommand{\bibinfo}[2]{#2}
\providecommand{\BIBentrySTDinterwordspacing}{\spaceskip=0pt\relax}
\providecommand{\BIBentryALTinterwordstretchfactor}{4}
\providecommand{\BIBentryALTinterwordspacing}{\spaceskip=\fontdimen2\font plus
\BIBentryALTinterwordstretchfactor\fontdimen3\font minus
  \fontdimen4\font\relax}
\providecommand{\BIBforeignlanguage}[2]{{%
\expandafter\ifx\csname l@#1\endcsname\relax
\typeout{** WARNING: IEEEtran.bst: No hyphenation pattern has been}%
\typeout{** loaded for the language `#1'. Using the pattern for}%
\typeout{** the default language instead.}%
\else
\language=\csname l@#1\endcsname
\fi
#2}}
\providecommand{\BIBdecl}{\relax}
\BIBdecl

\bibitem{Galceran2013}
E.~Galceran and M.~Carreras, ``A survey on coverage path planning for
  robotics,'' \emph{Robotics and Autonomous Systems}, vol.~61, no.~12, pp.
  1258--1276, 2013.

\bibitem{Almadhoun2019}
R.~Almadhoun, T.~Taha, L.~Seneviratne, and Y.~Zweiri, ``A survey on multi-robot
  coverage path planning for model reconstruction and mapping,'' \emph{SN
  Applied Sciences}, vol.~1, p. 847, 2019.

\bibitem{Bahamid2020}
A.~Bahamid, A.~Ibrahim, A.~Ibrahim, I.~Zahurin, and A.~Wahid, ``Intelligent
  robot-assisted evacuation: a review,'' \emph{Journal of Physics: Conference
  Series}, vol. 1706, no.~1, 2020.

\bibitem{Dhiman2021}
A.~Dhiman, N.~Shah, P.~Adhikari, S.~Kumbhar, I.~Dhanjal, and N.~Mehendale,
  ``Firefighting robot with deep learning and machine vision,'' \emph{Neural
  Computing and Applications}, vol.~34, pp. 2831--2839, 2021.

\bibitem{Feo1995GRASP}
T.~Feo and M.~Resende, ``Greedy randomized adaptive search procedures,''
  \emph{Journal of Global Optimization}, vol.~6, no.~2, pp. 109--133, Mar.
  1995.

\bibitem{12333}
F.~Yasutomi, M.~Yamada, and K.~Tsukamoto, ``Cleaning robot control,'' in
  \emph{International Conference on Robotics and Automation}, 1988, pp.
  1839--1841.

\bibitem{AbdRahman2022}
N.~Abd~Rahman, K.~Sahari, N.~Hamid, and Y.~Hou, ``A coverage path planning
  approach for autonomous radiation mapping with a mobile robot,''
  \emph{International Journal of Advanced Robotic Systems}, vol.~19, no.~4,
  2022.

\bibitem{Phung2017EnhancedDPSO}
M.~Phung, C.~Quach, T.~Dinh, and Q.~Ha, ``Enhanced discrete particle swarm
  optimization path planning for {UAV} vision-based surface inspection,''
  \emph{Automation in Construction}, vol.~81, pp. 25--33, 2017.

\bibitem{Becker2022}
K.~Becker, M.~Oehler, and O.~von Stryk, ``{3D} coverage path planning for
  efficient construction progress monitoring,'' in \emph{IEEE International
  Symposium on Safety, Security, and Rescue Robotics (SSRR)}, 2022.

\bibitem{Acar2002}
E.~Acar, H.~Choset, A.~Rizzi, P.~Atkar, and D.~Hull, ``Morse decompositions for
  coverage tasks,'' \emph{International Journal of Robotics Research}, vol.~21,
  no.~4, pp. 331--344, 2002.

\bibitem{LI2011876}
Y.~Li, H.~Chen, M.~{Joo Er}, and X.~Wang, ``Coverage path planning for {UAVs}
  based on enhanced exact cellular decomposition method,'' \emph{Mechatronics},
  vol.~21, no.~5, pp. 876--885, 2011, {S}pecial Issue on Development of
  Autonomous Unmanned Aerial Vehicles.

\bibitem{Oksanen2009}
T.~Oksanen and A.~Visala, ``Coverage path planning algorithms for agricultural
  field machines,'' \emph{Journal of Field Robotics}, vol.~26, no.~5, pp.
  651--668, 2009.

\bibitem{Choset1998}
H.~Choset and P.~Pignon, ``Coverage path planning: The boustrophedon cellular
  decomposition,'' in \emph{Field and Service Robotics}, A.~Zelinsky, Ed.\hskip
  1em plus 0.5em minus 0.4em\relax London: Springer London, 1998, pp. 203--209.

\bibitem{AcarChoset2002}
E.~Acar and H.~Choset, ``Sensor-based coverage of unknown environments:
  Incremental construction of morse decompositions,'' \emph{International
  Journal of Robotics Research}, vol.~21, no.~4, pp. 345--366, 2002.

\bibitem{Miao2018}
X.~Miao, J.~Lee, and B.-Y. Kang, ``Scalable coverage path planning for cleaning
  robots using rectangular map decomposition on large environments,''
  \emph{IEEE Access}, vol.~6, pp. 38\,200--38\,215, 2018.

\bibitem{Gabriely2001}
Y.~Gabriely and E.~Rimon, ``Spanning-tree-based coverage of continuous areas by
  a mobile robot,'' \emph{Annals of Mathematics and Artificial Intelligence},
  vol.~31, no.~1, pp. 77--98, 2001.

\bibitem{Swain2023}
S.~Swain, P.~Khilar, and B.~Senapati, ``An efficient path planning algorithm
  for {2D} ground area coverage using multi-{UAV},'' \emph{Wireless Personal
  Communication}, vol. 132, no.~1, pp. 361--407, Aug. 2023.

\bibitem{Mansouri2018}
S.~Mansouri, C.~Kanellakis, G.~Georgoulas, D.~Kominiak, T.~Gustafsson, and
  G.~Nikolakopoulos, ``{2D} visual area coverage and path planning coupled with
  camera footprints,'' \emph{Control Engineering Practice}, vol.~75, pp. 1--16,
  2018.

\bibitem{Ghaddar2020}
A.~Ghaddar, A.~Merei, and E.~Natalizio, ``{PPS}: Energy-aware grid-based
  coverage path planning for {UAVs} using area partitioning in the presence of
  {NFZs},'' \emph{Sensors}, vol.~20, p. 3742, 2020.

\bibitem{BarasDasygenis2023}
N.~Baras and M.~Dasygenis, ``{UGV} coverage path planning: An energy-efficient
  approach through turn reduction,'' \emph{Electronics}, vol.~12, p. 2959,
  2023.

\bibitem{TranGarrattKasmarikAnavatti2023}
V.~Tran, M.~Garratt, K.~Kasmarik, and S.~Anavatti, ``Dynamic frontier-led
  swarming: Multi-robot repeated coverage in dynamic environments,''
  \emph{IEEE/CAA Journal of Automatica Sinica}, vol.~10, no.~3, pp. 646--661,
  Mar. 2023.

\bibitem{TranGarrattKasmarikAnavattiAbpeikar2022}
V.~Tran, M.~Garratt, K.~Kasmarik, S.~Anavatti, and S.~Abpeikar, ``Frontier-led
  swarming: Robust multi-robot coverage of unknown environments,'' \emph{Swarm
  and Evolutionary Computation}, vol.~75, p. 101171, 2022.

\bibitem{KantarosThanouTzes2015}
Y.~Kantaros, M.~Thanou, and A.~Tzes, ``Distributed coverage control for concave
  areas by a heterogeneous robot–swarm with visibility sensing constraints,''
  \emph{Automatica}, vol.~53, pp. 195--207, 2015.

\bibitem{Tran2023multipleugvs}
V.~Tran, A.~Perera, M.~Garratt, K.~Kasmarik, and S.~Anavatti, ``Coverage path
  planning with budget constraints for multiple unmanned ground vehicles,''
  \emph{IEEE Transactions on Intelligent Transportation Systems}, vol.~24,
  no.~11, pp. 12\,506--12\,522, 2023.

\bibitem{An2018}
V.~An, Z.~Qu, and R.~Roberts, ``A rainbow coverage path planning for a
  patrolling mobile robot with circular sensing range,'' \emph{IEEE
  Transactions on Systems, Man, and Cybernetics: Systems}, vol.~48, no.~8, pp.
  1238--1254, 2018.

\bibitem{Jonnarth2023}
\BIBentryALTinterwordspacing
A.~Jonnarth, J.~Zhao, and M.~Felsberg, ``Learning coverage paths in unknown
  environments with deep reinforcement learning,'' \emph{arXiv preprint}, 2023.
  [Online]. Available: \url{https://arxiv.org/abs/2306.16978}
\BIBentrySTDinterwordspacing

\bibitem{FernandezTalavera2023}
N.~{Fern{\'a}ndez Talavera}, J.~Rold{\'a}n-G{\'o}mez, F.~Mart{\'i}n, and
  M.~Rodriguez-Sanchez, ``An autonomous ground robot to support firefighters'
  interventions in indoor emergencies,'' \emph{Journal of Field Robotics},
  vol.~40, no.~3, pp. 451--473, May 2023.

\bibitem{Nasirian2021}
B.~Nasirian, M.~Mehrandezh, and F.~Janabi-Sharifi, ``Efficient coverage path
  planning for mobile disinfecting robots using a graph-based representation of
  environment,'' \emph{Frontiers in Robotics and AI}, vol.~8, 2021.

\bibitem{Pierson2021}
A.~Pierson, J.~Romanishin, H.~Hansen, L.~Yanez, and D.~Rus, ``Designing and
  deploying a mobile {UVC} disinfection robot,'' in \emph{International
  Conference on Intelligent Robots and Systems}, 2021.

\bibitem{perminov2023}
S.~Perminov, I.~Kalinov, and D.~Tsetserukou, ``{GHACPP}: Genetic-based
  human-aware coverage path planning algorithm for autonomous disinfection
  robot,'' in \emph{International Conference on Systems, Man, and Cybernetics},
  2023.

\bibitem{Chen2024}
Z.~Chen, H.~Wang, K.~Chen, C.~Song, X.~Zhang, B.~Wang, and J.~C. Cheng,
  ``Improved coverage path planning for indoor robots based on bim and robotic
  configurations,'' \emph{Automation in Construction}, vol. 144, p. 105160,
  2024.

\bibitem{Chen2022}
Z.~Chen, K.~Chen, C.~Song, X.~Zhang, J.~Cheng, and D.~Li, ``Global path
  planning based on {BIM} and physics engine for {UGVs} in indoor
  environments,'' \emph{Automation in Construction}, vol. 139, p. 104263, 2022.

\bibitem{Lee2023}
J.-K. Lee, A.~Naser, O.~Ennasr, A.~Soylemezoglu, and G.~Glaspell, ``Unmanned
  ground vehicle ({UGV}) full coverage planning with negative obstacles,''
  Engineer Research and Development Center (U.S.), Tech. Rep. ERDC TR-23-13,
  2023.

\bibitem{FraunhoferIPA2016}
R.~Bormann, F.~Jordan, W.~Li, J.~Hampp, and M.~Haegele, ``Room segmentation:
  Survey, implementation, and analysis,'' in \emph{International Conference on
  Robotics and Automation}, 2016.

\bibitem{FraunhoferIPA2018}
R.~Bormann, F.~Jordan, J.~Hampp, and M.~Haegele, ``Indoor coverage path
  planning: Survey, implementation, analysis,'' in \emph{International
  Conference on Robotics and Automation}, 2018.

\bibitem{bresenham1965}
J.~Bresenham, ``Algorithm for computer control of a digital plotter,''
  \emph{IBM Systems journal}, vol.~4, no.~1, pp. 25--30, 1965.

\bibitem{hart1968astar}
P.~Hart, N.~Nilsson, and B.~Raphael, ``A formal basis for the heuristic
  determination of minimum cost paths,'' \emph{IEEE Transactions on Systems
  Science and Cybernetics}, vol.~4, no.~2, pp. 100--107, 1968.

\bibitem{gmapping2007}
G.~Grisetti, C.~Stachniss, and W.~Burgard, ``Improved techniques for grid
  mapping with rao-blackwellized particle filters,'' \emph{IEEE Transactions on
  Robotics}, vol.~23, no.~1, pp. 34--46, 2007.

\bibitem{Yang2004}
S.~Yang and C.~Luo, ``A neural network approach to complete coverage path
  planning,'' \emph{IEEE Transactions on Systems, Man, and Cybernetics, Part B
  (Cybernetics)}, vol.~34, no.~1, 2004.

\bibitem{Arain2015}
M.~Arain, M.~Cirillo, V.~Bennetts, E.~Schaffernicht, M.~Trincavelli, and
  A.~Lilienthal, ``Efficient measurement planning for remote gas sensing with
  mobile robots,'' in \emph{International Conference on Robotics and
  Automation}, 2015.

\bibitem{Bormann2015}
R.~Bormann, J.~Hampp, and M.~H{\"a}gele, ``New brooms sweep clean - an
  autonomous robotic cleaning assistant for professional office cleaning,'' in
  \emph{International Conference on Robotics and Automation}, 2015.

\end{thebibliography}

\end{document}